# ARN-LSTM: A Multi-Stream Fusion Model for Skeleton-based Action Recognition

Chuanchuan WANG[1], Ahmad Sufril Azlan Mohmamed[1,*], Mohd Halim Bin Mohd Noor[1], Xiao Yang[1], Feifan Yi[1,3], Xiang Li[2]
1 School of Computer Science, Universiti Sains Malaysia, Penang, 11800, Malaysia
2 School of Engineering, College of Technology and Business, Guangzhou, 510850, China
3 Institute of Applied Artificial Intelligence of the Guangdong-Hong Kong-Macao Greater Bay Area, Shenzhen Polytechnic University, 518102, China
*Corresponding author: Ahmad Sufril Azlan Mohamed (sufril@usm.my)

Abstract: This paper presents the ARN-LSTM architecture, a novel multi-stream action recognition model designed to address the challenge of simultaneously capturing spatial motion and temporal dynamics in action sequences. Traditional methods often focus solely on spatial or temporal features, limiting their ability to comprehend complex human activities fully. Our proposed model integrates joint, motion, and temporal information through a multi-stream fusion architecture. Specifically, it comprises a joint stream for extracting skeleton features, a temporal stream for capturing dynamic temporal features, and an ARN-LSTM block that utilizes Time-Distributed Long Short-Term Memory (TD-LSTM) layers followed by an Attention Relation Network (ARN) to model temporal relations. The outputs from these streams are fused in a fully connected layer to provide the final action prediction. Evaluations on the NTU RGB+D 60 and NTU RGB+D 120 datasets outperform the superior performance of our model, particularly in group activity recognition.

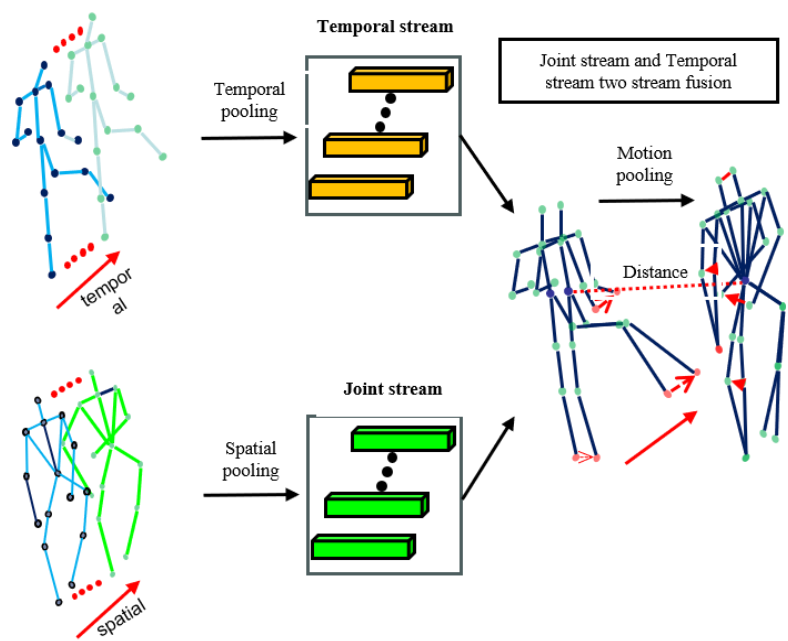

Figure 1 The overview of the proposed method.

## 1. Introduction

Action recognition is a key task in computer vision, with applications spanning human-computer interaction, video surveillance, healthcare, and robotics[1-4]. Recent advancements have significantly improved the accuracy of action recognition models, yet challenges remain, particularly in effectively capturing both spatial and temporal dynamics in human activity sequences. Human action recognition methods can be classified into two primary categories: RGB-based and skeleton-based[5, 6]. The latter extracts skeleton points from input videos, modeling their feature changes across frames to facilitate action recognition. Compared to RGB-based methods[7, 8], these approaches lower hardware demands and computational costs, while being less vulnerable to interference from extraneous factors such as lighting or background[9]. Traditional methods either focus on spatial or temporal information in isolation, limiting their ability to understand complex activities involving motion and temporal patterns fully. Traditional skeleton-based interaction recognition approaches treat human joints as independent features, constructing feature sequences from them for input into recurrent or convolutional neural networks to predict actions. However, these methods fail to account for the correlation features between different joints[10].

Most existing single-person[10, 11] action recognition methods have adopted the concept of graph convolution. However, in action recognition, scenarios often involve both dual-person interactions and single-person movements. Traditional graph convolution models typically treat the two individuals in dual-person interactions as isolated entities, thereby neglecting critical interaction information between them.

Therefore, to effectively extract the interaction information between behaviors and be inspired by the effectiveness of the feature fusion ide[12-14], we propose the ARN-LSTM architecture for human action recognition. Figure *1* shows, that the ARN-LSTM approach is a multi-stream action recognition model designed to fuse joint, motion, and temporal features to address these challenges. Our model leverages Attention Relation Networks (ARN)[15] to enhance the correlations between these features, allowing for more comprehensive action recognition. By combining joint skeletal data with temporal dynamics, ARN-LSTM provides a more complete understanding of human activity, this fusion of streams allows ARN-LSTM to handle more complex action sequences, particularly in group activity recognition tasks.

The main contributions of this work are summarized as follows:

1. We proposed a multi-stream architecture that fuses joint, motion, and temporal information for improved action recognition, significantly enhancing action recognition capabilities. It serves to perform fusion among multiple relational models and accommodates various fusion types (e.g., separate individual models, inter-relational models, etc.). This approach flexibly integrates diverse forms of relational information to construct the final input feature representation.

2. This work introduces the Attention Relation Network (ARN), a novel mechanism designed to effectively capture and amplify the interrelationships between spatial and temporal features, thereby refining the model's analytical prowess.

3. Our proposed ARN-LSTM model achieves state-of-the-art performance on the NTU RGB+D 60/120 dataset, demonstrating its robust capacity to manage and interpret complex group activities with high precision.

These contributions culminate in the innovative architecture ARN-LSTM, which is both straightforward and effective. We conduct experiments on the largest and most widely used action-recognition dataset encompassing human interactions across varied conditions, achieving state-of-the-art performance and competitive results that underscore the robustness of our proposed approach.

The remainder of this paper is organized as follows: Section 2 discusses related work in action recognition. Section 3 Explains our proposed ARN-LSTM model, including its multi-stream architecture, the TD-LSTM architecture and fusion mechanism. Section 4 outlines the experimental setup, including the dataset, evaluation metrics, and implementation details. Section 5 presents the results and discusses the model's performance, followed by a conclusion in Section 6.

## 2. Related work

Skeleton-based human action recognition became a hotspot in the field of computer vision, about of skeleton dataset emerged and it was witnessed that a novel architecture, such as GCN-based[16], CNN-based[17] RNN-based and LSTM-based[18]. Most have beaten many established baselines in several applications. The ST-GCN[19] models the skeleton data as a graph structure under the spatio-temporal GCN architecture. Spatio-temporal graph Convolutional Networks marries the spatial graph convolutions with the temporal convolutions for modeling the spatio-temporal sequence data, becoming the de facto model in the paradigm. Upon the baseline, many variants of ST-GCN have been proposed to enhance the modeling capacity. For example, [17]investigated Shift-GCN which employs shift graph operations and lightweight point-wise convolutions to alleviate the computational complexity burden. Ref. [20] proposed a channel-wise topology refinement graph convolution to dynamically model different topologies, via aggregating joint features in different channels of skeletons. The InfoGCN[21] architecture combined with an Information-Bottleneck learning objective and a self-attention-based graph convolution module, learns the compressed latent representation of actions and infers context-dependent intrinsic topology in spatial modelling of skeletons. They also suggested a hierarchically decomposed GCN[22] that extracts major structural edges and uses them to construct a hierarchically decomposed graph. Other follow-up works include multi-scale modelling[23], graph routing[24, 25].

Action recognition has seen considerable advancements with the rise of deep learning techniques, particularly convolutional neural networks (CNNs) and recurrent neural networks (RNNs)[26]. Traditional methods can be broadly classified into two categories: spatial-based, temporal-based, and spatial-temporal-based approaches. This section will briefly review the related works about the three fields of skeleton-based action recognition.

### 2.1 Spatial-based Approaches

Spatial-based methods capture spatial features such as human poses, body parts, and skeletal structures. Early works primarily relied on hand-crafted features, which were limited in capturing complex motion patterns. With the advent of deep learning, CNN-based[27] approaches have emerged, significantly improving the ability to extract spatial features from video frames. Skeleton-based approaches, such as those that extract joint positions, have also gained popularity due to their robustness against variations in appearance and lighting.

## 2.2 Temporal-based Approaches

Temporal-based methods, on the other hand, focus on modeling the temporal evolution of actions over time. Long Short-Term Memory (LSTM) networks and Temporal Convolutional Networks (TCNs) have widely captured temporal dependencies in action sequences. These models have demonstrated success in handling the sequential nature of action data, yet they often struggle with long-range temporal dependencies and complex temporal dynamics. Early deep learning approaches in human activity recognition (HAR) typically employed CNN[7, 28] or RNN[10, 29, 30] for temporal modeling. For example, Du et al.[30] applied 3D convolution to sequence segments to model temporal dynamics, pooling along the temporal dimension for the final feature representation. Another common approach involves converting sequences into 2D representations, with 2DCNNs[31] used to capture temporal dynamics. In studies such as[32] discrete joint trajectories were color-encoded to create 2D images, referred to as Joint Trajectory Maps, which were then processed using CNN-based models to extract temporal features. However, representing skeleton sequences as pseudo-images fails to fully capture the complex interconnectivity among human joints.

Recent models [33-36]employ temporal convolutions at three fixed scales[33] to capture long, medium, and short-term temporal graph convolution network characteristics of individual joints. This was further enhanced by incorporating spatially neighboring joints[37] when applying temporal convolutions. Zhifuetal.[38] extended this approach by incorporating predefined body parts, though such predefined parts may inevitably include non-informative joints. In[39],sample-dependent weights were introduced during scale fusion. Notably, all of these multi-scale temporal methods rely on conventional average or max pooling for feature aggregation[33, 39].

## 2.3 Multi-stream Fusion Models

Multi-stream architectures that integrate both spatial and temporal features have shown promise in overcoming the limitations of single-stream models[40, 41]. Two-stream networks[19], for example, process spatial and temporal information in parallel using separate CNNs for RGB frames and optical flow, and later fuse the outputs[42]. Although these models provide improved accuracy, they still suffer from limited fusion mechanisms that do not fully capture the interdependencies between spatial and temporal features, such as the motion interaction relation information. [43] created a graph convolutional LSTM to isolate distinct spatial and temporal characteristics from video content. Dai and colleagues [44] suggested a dual-stream attention-oriented LSTM for recognizing actions, capable of focusing on distinguishing features to enhance performance. Furthermore, Ma and colleagues. [45] suggested the development of a temporal segment LSTM to encapsulate correlations by integrating spatial and temporal attributes into feature matrices within the temporal realm.

However, the earlier models based on RNNs overlooked the integration of knowledge from various streams, and multi-feature fusion mainly focuses on video-based human action recognition[46]. Diverging from earlier models based on two-stream CNNs and RNNs, our ARN-LSTM model uniquely captures motion information and leverages the synergistic insights from various streams through multi-task learning, such as joint stream, temporal stream, and multi-stream fusion features. Overall, existing methods face limitations in the simultaneous selection of discriminative frames and joints during temporal pooling. Furthermore, they struggle to model the long-range cross-joint relationships among informative joints effectively. The proposed ARN-LSTM aims to address these challenges. Our ARN-LSTM model builds on this idea of multi-stream fusion but introduces a more sophisticated fusion mechanism using Attention Relation Networks, enabling a deeper integration of spatial and temporal features.

## 3 Proposed ARN-LSTM method

In this section, we describe the architecture of the ARN-LSTM model and its key components. Previous research indicates that the simultaneous use of different streams can significantly improve the performance of human action recognition[51][47]. Consequently, we assess the performance of the trained models employing streams for joint, bone, joint motion, and bone motion. The bone stream uses bone modality as input data, as proposed by Shi et al[48]. The joint motion and bone motion streams are consistent with the methodology outlined by Shi et al.[47]. The final result is determined by calculating a weighted average of the inference outputs from the models.

## 3.1 Overview

Inspired by the RN[49], IRN[54][50], and ARN[15] architectures, we designed an ARN-LSTM method. The architecture embodies three types of interaction relationships: inward-person relationships, outward-person relationships, and the fusion of the two features.

Definea sophisticated neural network architecture designed to process inputs involving diverse objects (e.g., human body joint data) while employing relational modeling and attention mechanisms to enhance model performance. This architecture is particularly suited for handling complex data characterized by multiple inputs and relationships, especially in tasks such as behavior recognition or human posture estimation. With its flexible parameter configurations, users can adapt the model structure to fulfill specific task requirements.

In the selection of weights during the relational modeling process, a Gaussian[51, 52] function was employed to generate weights that prioritize the central sequence. The Gauss formula is shown in (1):

$$gauss(x) = \frac{1}{\sigma\sqrt{2\pi}} \exp\left(-\frac{x^2}{2\sigma^2}\right) \quad (1)$$

For the inward-person relation use as $ARN - LSTM_{inward}(P_1, P_2)$ corresponding to equations (2):

$$ARN - LSTM_{inward}(P_1, P_2) = f_\emptyset\left(\sum_{i,k} g_\theta(j_i^1, j_k^2) \oplus \sum_{i,k} g_\theta(j_i^2, j_k^1)\right) \quad (2)$$

In theory, fφ and gθ can represent Multi-Layer Perceptrons (MLPs) characterized by trainable parameters φ and θ, respectively[16]. Notably, it ⊕ can encompass various pooling operations, including summation, maximization, averaging, or concatenation[54]. However, based on our experimental findings, we have opted to employ the averaging operation due to its superior performance.

For the outward-person relation is $ARN - LSTM_{outward}(P_1, P_2)$ corresponding to equations (3):

$$ARN - LSTM_{outward}(P_1, P_2) = f_{\emptyset'}\left(\sum_{i=1}^{N}\sum_{K=i+1}^{N} g_\Theta(j_i^1, j_k^1)\right.$$

$$\left.\frown \sum_{i=1}^{N}\sum_{K=i+1}^{N} g_\Theta(j_i^2, j_k^2)\right) \quad (3)$$

Recognizing that intra-personal relationships among joints can provide critical insights, we propose an innovative architecture wherein the joints of each individual are paired exclusively with their corresponding joints. In this framework, bidirectional pairing is superfluous since the joint pairs stem from a single individual. Our preliminary experiments indicate that incorporating bidirectional pairing may introduce redundant complexity into the model, potentially leading to overfitting in certain cases. This finding underscores the necessity for a more streamlined approach to joint pairing that retains essential structural relationships while minimizing extraneous redundancy. By focusing on the inherent relationships within an individual's configuration, our architecture aims to enhance model efficiency and generalization. This streamlined design not only simplifies the computational requirements but also fortifies the robustness of the recognition process. Our observations suggest that eliminating unnecessary bidirectional connections could pave the way for improved accuracy and interpretability in human action recognition tasks. Thus, we advocate for a paradigm shift towards leveraging intra-personal joint relationships as a foundational element in the design of action recognition architectures, with the potential to yield significant advancements in the field. The aggregated output from each individual is concatenated (⌢) before being processed through function f, characterized by its trainable parameters φ.

$ARN - LSTM_{inward+outward}(P_1, P_2)$ rrepresents for the inward-person and outward-person fusion relation, corresponding to equations (4):

$$ARN - LSTM_{inward+outward}(P_1, P_2) = f_{\emptyset''}\left(\sum_{i,k} g_\theta(j_i^1, j_k^2) \oplus \sum_{i,k} g_\theta(j_i^2, j_k^1) \sum_{i=1}^{N}\sum_{K=i+1}^{N} g_\Theta(j_i^1, j_k^1)\right.$$

$$\left.\frown \sum_{i=1}^{N}\sum_{K=i+1}^{N} g_\Theta(j_i^2, j_k^2)\right) \quad (4)$$

The relationship modeling module constructs distinct relationship matrices according to the specific type of relationship (e.g., intra-monadic relationships, inter-individual relationships) and processes each pair of input objects through the gθ network. Conclusively, we propose an architecture that amalgamates both categories of relationships within a unified function f (parametrically defined by φ), achieved through concatenating the pooled information from each function g, each governed by its distinct parameters φ and θ.

The ARN-LSTM model is designed to capture both spatial and temporal features in action recognition tasks. The model comprises three primary components: the joint stream, the temporal Stream, and the ARN-LSTM block(the ARN method with LSTM layers). Each stream processes distinct aspects of the input data, which are then fused to generate the final action prediction[15].

(1) The joint stream. The Joint stream focuses on extracting spatial features from the human skeleton. By leveraging skeletal data, the model captures spatial configurations of human joints, critical for identifying poses and movements. In this paper, leveraging sampled skeleton frames as input, the spatial stream gathers worldwide spatial data from the videos. In the case of this stream, our initial step involves sampling several skeleton frames periodically and sequentially feeding them into the network. Subsequently, we calculate the average of the spatial stream's losses across all chosen sampled RGB video frames to determine the final loss of the spatial stream[53]. The output predicted by the spatial stream is labeled as y_joint.

(2) The temporal stream. The Temporal stream captures temporal dynamics in the input sequence. This stream models the evolution of joint movements over time, providing insights into the dynamic nature of actions. The temporal stream captures the global motion information of each pixel by using the sampled stacked optical flow frames as the input. In this paper, for the temporal stream, as the 3M approach[53], our architectural also sample 2L stacked optical flow frames including L stacked optical flow frames in the vertical direction and L stacked optical flow in the horizontal direction, feed the sampled stacked optical flow frames into the temporal stream to obtain the output of the temporal stream denoted as y_temporal [53].

(3) The Time distributed Long short-term memory block. The ARN-LSTM block is the model's core and captures both short and long-term temporal dependencies. It consists of Temporal-Deep Long Short-Term Memory (TD-LSTM) layers and an attention relation network[15]. The TD-LSTM layers model temporal relationships within each stream, while the ARN enhances the correlations between the joint and temporal features, allowing the model to focus on the most relevant aspects of the action sequence.

### 3.2 TD-LSTM

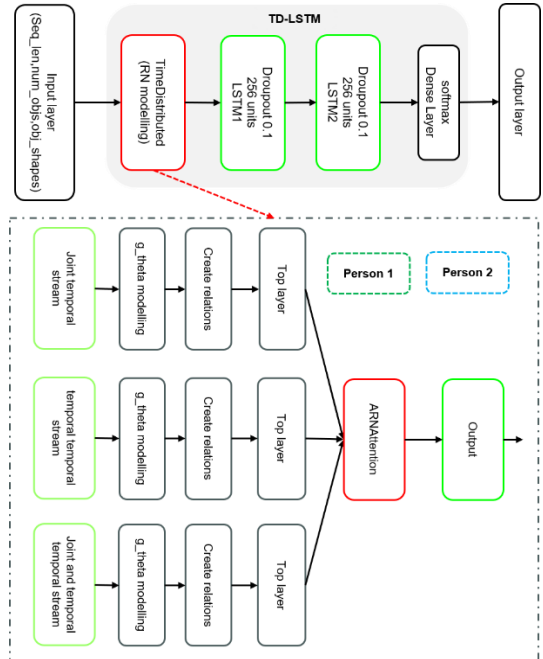

Figure 2 Illustration of the proposed Time Distributed LSTM architecture

The Time Distributed LSTM (TD-LSTM) comprises the main components depicted in Figure 2. The input layer consists of several objects, each with a specific object shape dimension. The distributed layer applies the same neural network across the time dimension, while the LSTM component includes two separate LSTM layers for time series processing. The dense layer, serving as the output layer, utilizes the softmax activation function to generate the final output. Note that in the Time distributed layer, the person-to-person relationship model needs to be taken into account, and the input sequences include joint stream with motion, temporal stream with motion, and joint fused temporal with motion, separately. For a detailed insight into the TD-LSTM method flow, please refer to Algorithm 1.

**Algorithm 1** The LSTM Integration with Temporal Distributed Model

**Input:**
- object_shape: Shape of each object
- seq_len: Sequence length

- prune_at_layer: Pruning layer
- kernel_init: Kernel initializer
- g_theta_kwargs:
- drop_rate: Dropout rate

**Output:** The Compiled Temporal Relation Network model

1: g_theta_model ← initialize_g_theta(object_shape, kernel_init, **g_theta_kwargs)
2: **if** prune_at_layer is not None:
3:   **for** layer **in** reversed(g_theta_model.layers):
4:     **if** layer_name ends with prune_at_layer:
5:       top_layer_out ← layer.output
6:     **break**
7:   g_theta_model ← Model(inputs=g_theta_model.input, outputs=top_layer_out)
8: input_g_theta ← create_input(shape=((2,)+object_shape))
9: g_theta_model_out ← g_theta_model(Lambda(slice_input)(input_g_theta))
10: merged_g_theta_model ← Model(inputs=input_g_theta, outputs=g_theta_model_out)
11: temporal_input ← create_input(shape=((seq_len, 2,) + object_shape))
12: x ← TimeDistributed(merged_g_theta_model)(temporal_input)
13: x ← LSTM(500, dropout=drop_rate, return_sequences=True)(x)
14: g_theta_lstm_model ← Model(inputs=temporal_input, outputs=x, name='g_theta_lstm')
15: **return** g_theta_lstm_model

3.3 Multi-stream Fusion

After processing the input data through the joint stream and temporal streams, the outputs are fused using a fully connected layer. This fusion layer combines the spatial and temporal features, enabling the model to represent the action comprehensively. The fused features are then passed through a final softmax layer to produce the action prediction.

This approach constructs a sophisticated neural network model adept at processing diverse forms of object data, including federated streams, joint streams, temporal streams, and general relational data. It leverages relational modeling and attention mechanisms to enhance model performance. The flexible parameter configuration facilitates the implementation of various network architectures.

4 Experimental

In this work, the evaluation of the proposed architecture was run on an Ubuntu12.04 operation system with a 24GB memory NVIDIA GeForce graphics card.

4.1 Dataset

The ARN-LSTM model was evaluated on the NTU RGB+D 60[54] and NTU RGB+D 120[55] datasets. The NTU RGB+D dataset is the largest and most widely used action-recognition dataset.

NTU RGB+D 60 dataset[54]. The NTU RGB+D 60 dataset comprises 56,880 skeleton action sequences across 60 classes. Each sample represents a single action, performed by up to two subjects and recorded by three cameras from varying perspectives. The dataset is partitioned into two test benchmarks based on distinct subjects and views: cross-subject (or X-Sub) and cross-view (or X-View)[54].

NTU RGB+D 120 dataset[55]. The NTU RGB+D 120 dataset extends the NTU RGB+D collection, comprising 114,480 samples across 120 classes. This dataset was captured using three cameras and encompasses 32 settings, each representing a specific location and background. It is organized into two benchmarks based on subject parity and sample IDs: cross-subject (or X-Sub) and cross-setup (or X-View)[55]. The dataset includes both RGB frames and 3D skeleton data, making it ideal for evaluating the performance of multi-stream models like ARN-LSTM.

In this work, the skeletons should generate a single CSV file with all the normalized coordinates in this work. The datasets used or analyzed during the current study are publicly available from the link https://doi.org/10.6084/m9.figshare. 27427188.v1. 26 mutual actions were used for evaluation, and Openpose extracted four sample skeleton-based inter-action actions, shown in Figure 3.

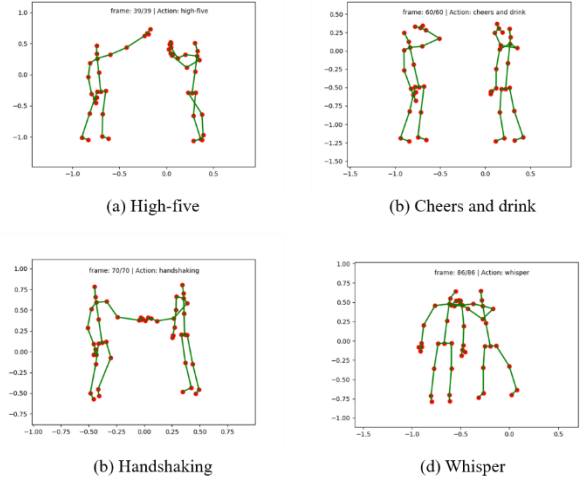

Figure 3 Sample skeleton actions of the NTU RGB+D dataset.(a) high-five,(b)cheers and drink,(c)handshaking and(d)whisper.

4.2 Evaluation Metrics

We used standard metrics for action recognition evaluation, including accuracy, precision, recall, and F1-score[56]. The confusion matrix was also analyzed to assess the model's robustness in differentiating between similar action classes. For the binary classification recognition problem, a confusion matrix is used to evaluate performance on test data shown in Figure 4, the Accuracy, Precision, Recall and F-Score equation as described in (5),(6),(7),(8) separately.

$$Accuracy = \frac{TP+TN}{TP+TN+FP+FN} \quad (5)$$

$$Pecision = \frac{TP}{TP+FP} \quad (6)$$

$$Recall = \frac{TP}{TP+FN} \quad (7)$$

$$F - Score = 2 * \frac{Precesion*Recall}{Precesion+Recall} \quad (8)$$

4.3 Implementation Details

The ARN-LSTM model was implemented using Tensorflow 2.8.0 and Python 3.8.16. The Joint and Temporal streams were initialized without pre-trained weights, and the ARN-LSTM block was trained from scratch. The model was optimized using Adam with a learning rate of 0.0001. The training was conducted on a single NVIDIA RTX4090 GPU, with a batch size of 64, a learning rate of 0.0001, a drop rate of 0.1 and early stopping based on validation accuracy.

4.4 Comparison of multi-stream feature

We experiment with the proposed approach of multi-stream feature comparisons, acknowledging that coordinates are not wholly accurate and often contain noise along with occasional tracking errors. To more precisely reflect the algorithm's performance, we employ a 5-fold cross-validation protocol to report each fold's accuracy of the ARN-LSTM approach. Table 1 presents a performance comparison of ARN-LSTM models with different multi-stream fusion features (joint, temporal, and motion) on the NTU RGB+D 60 dataset, evaluated on both cross-subject and cross-view tasks.

Table 1 Comparison of ARN-LSTM with multi-stream feature (containing joint, temporal and motion) methods on the NTU RGB+D 60 dataset.

| Methods | Fold no. | Cross-subject | | Cross-view | |
|---|---|---|---|---|---|
| | | Acc(%) | Loss | Acc(%) | Loss |
| joint+motion | 0 | 92.46 | 0.20 | 95.30 | 0.13 |
| | 1 | 95.74 | 0.13 | 91.57 | 0.24 |
| | 2 | 94.85 | 0.15 | 93.05 | 0.19 |
| | 3 | 95.66 | 0.13 | 93.57 | 0.18 |
| | 4 | 92.99 | 0.20 | 95.41 | 0.12 |
| temporal+motion | 0 | 91.80 | 0.22 | 94.57 | 0.15 |
| | 1 | 95.22 | 0.13 | 92.89 | 0.19 |
| | 2 | 92.95 | 0.19 | 92.92 | 0.19 |
| | 3 | 93.82 | 0.16 | 93.64 | 0.18 |

| Methods | Fold no. | | | | |
|---|---|---|---|---|---|
| | 4 | 93.44 | 0.17 | 94.93 | 0.14 |
| | 0 | 91.16 | 0.24 | **98.62** | **0.05** |
| | 1 | **98.80** | **0.03** | 98.58 | 0.04 |
| Joint+temporal+motion | 2 | 97.99 | 0.06 | 97.60 | 0.07 |
| | 3 | 98.50 | 0.04 | 96.11 | 0.11 |
| | 4 | 95.86 | 0.11 | 98.03 | 0.06 |

Three configurations are compared:

**Joint with motion:** For cross-subject, accuracy ranges from 92.46% to 95.74% with a loss between 0.13 and 0.20. For cross-view, accuracy ranges from 91.57% to 95.41% with a loss between 0.12 and 0.24.

Temporal with Motion: For cross-subject, accuracy ranges from 91.80% to 95.22% with a loss between 0.13 and 0.22. For cross-view, accuracy ranges from 92.89% to 94.93% with a loss between 0.14 and 0.19.

**Joint fusion temporal with motion:** For cross-subject, accuracy ranges from 91. 16% to 98.80% with a loss between 0.03 and 0.24. For cross-view, accuracy ranges from 96. 11% to 98.62% with a loss between 0.04 and 0.11.

The Joint fusion temporal with motion configuration achieves the best performance, especially in cross-view tasks, where accuracy reaches 98.62%.

Table 2 Comparison of ARN-LSTM with multi-stream feature (containing joint, temporal and motion information) methods on the NTU RGB+D 120 dataset.

| Methods | Fold no. | Cross-subject(%) | | Cross-view(%) | |
|---|---|---|---|---|---|
| | | Acc(%) | Loss | Acc(%) | Loss |
| | 0 | 92.01 | 0.23 | 91.97 | 0.23 |
| | 1 | 93.04 | 0.20 | 91.66 | 0.24 |
| joint+motion | 2 | 86.12 | 0.10 | 93.34 | 0.20 |
| | 3 | 93.92 | 0.17 | 91.89 | 0.24 |
| | 4 | 91.44 | 0.25 | 93.90 | 0.18 |
| | 0 | 91.39 | 0.24 | 92.14 | 0.22 |
| | 1 | 91.36 | 0.24 | 92.10 | 0.22 |
| temporal+motion | 2 | 92.46 | 0.21 | 89.62 | 0.29 |
| | 3 | 90.07 | 0.28 | 91.72 | 0.23 |
| | 4 | 92.56 | 0.20 | 91.48 | 0.23 |
| | 0 | 98.74 | 0.04 | 96.70 | 0.10 |
| | 1 | 97.95 | 0.06 | 99.31 | 0.02 |
| Joint+temporal+motion | 2 | **99.47** | **0.02** | 99.22 | 0.03 |
| | 3 | 98.90 | 0.04 | **99.49** | **0.02** |
| | 4 | 97.87 | 0.07 | 99.09 | 0.03 |

Table 2 provides a comparative analysis of ARN-LSTM using various multi-stream features—specifically joint, temporal, and motion information—evaluated on the NTU RGB+D 120 dataset under cross-subject and cross-view settings. It demonstrates that incorporating all three feature streams (joint + temporal + motion) consistently yields the highest accuracy and the lowest loss across different folds. Notably, the joint+ temporal + motion configuration achieves peak accuracy of 99.47% in the cross-subject setting and 99.49% in the cross-view setting, with minimal associated loss values (0.02), indicating a significant performance advantage over configurations that combine only two of the streams. This highlights the efficacy of integrating comprehensive multi-stream data for enhanced action recognition performance in diverse viewpoints and subject variations.

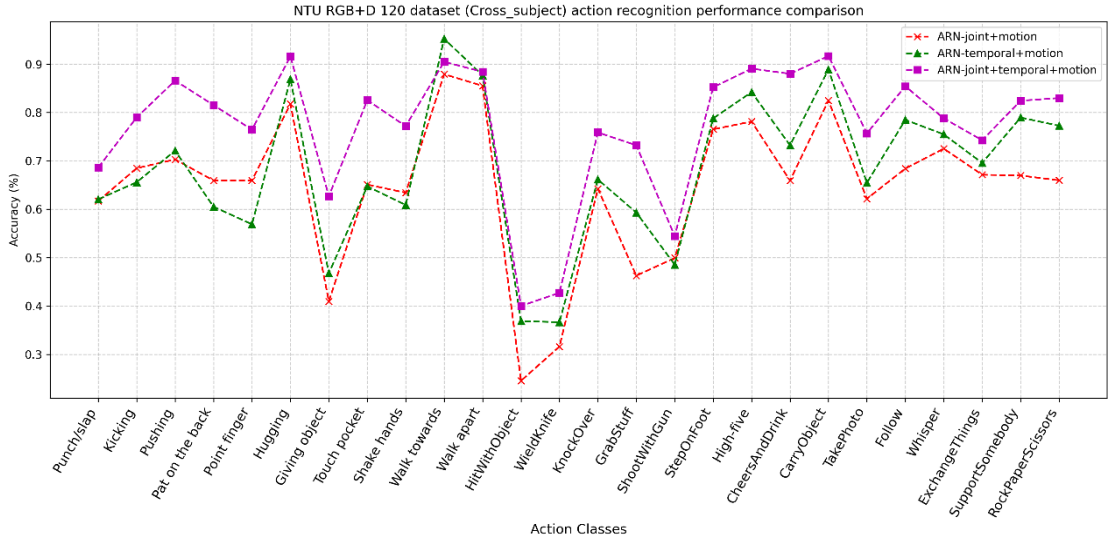

Figure 4 Different stream feature method's accuracy comparison in the NTU RGB+D 120 dataset.

Figure 4 presents action recognition accuracy comparisons for the NTU RGB+D 120 dataset (cross-subject) across various action classes using three fusion models: ARN-joint+motion, ARN-temporal+motion, and ARN-joint+temporal+motion. The ARN-joint+temporal+motion model (purple) consistently achieves higher accuracy across most action classes, notably outperforming the other models in actions such as "Hugging," "Shake hands," and "Wield knife" In general, the ARN-temporal+motion model (green) performs better than the ARN-joint+motion model (red) in several action classes, though the performance gap varies across different actions.

In summary, the two graphs compare action recognition performance on the NTU RGB+D 120 dataset (cross-view and cross-subject) across three fusion models, with ARN-joint+temporal+motion consistently outperforming ARN-joint+motion and ARN-temporal+motion across most action classes.

## 5 Results and Discussion
### 5.1 Performance Comparison

Experiment performance compared with the previous technical, to more precisely reflect the algorithm's superior performance, we leverage a 5-fold cross-validation protocol to report the top-1 accuracy of our approach.

Table 3 Compare with the previous method on the NTU RGB + D 60/120 dataset.

| Methods | Year | NTU RGB+D 60 | | NTU RGB+D 120 | |
|---|---|---|---|---|---|
| | | Cross-setup(%) | Cross-view(%) | Cross-setup(%) | Cross-view(%) |
| ST-LSTM[57] | 2016 | 69.2 | 77.7 | 55.7 | 57.9 |
| STA-LSTM[58] | 2017 | 73.4 | 81.2 | -- | -- |
| ST-GCN[19] | 2018 | 59.1 | 64.0 | 70.7 | 73.2 |
| 2s-AGCN[48] | 2019 | 88.5 | 95.1 | 82.5 | 84.2 |
| 4s-shift-GCN[17] | 2020 | 90.7 | 96.5 | 85.9 | 87.6 |
| MCC+2s-AGCN[59] | 2021 | 89.7 | 96.3 | 81.3 | 83.3 |
| EfficentGCN[60] | 2022 | 92.1 | 96.1 | 88.7 | 88.9 |
| PA-GCN[16] | 2023 | 92.1 | 96.7 | 87.4 | 89.8 |
| ARN*[15] | 2023 | 93.0 | 93.6 | 92.1 | 93.3 |
| ARN-LSTM_inward | -- | 97.1 | 97.2 | 94.9 | 91.8 |
| ARN-LSTM_outward | -- | 94.5 | 98.4 | 97.7 | 94.3 |
| ARN-LSTM_inward+outward | -- | **97.1** | **99.4** | **99.7** | **95.8** |

\* represents the results obtained from the reproduction of the paper.

Table 3 summarizes the performance of ARN-LSTM compared to other state-of-the-art methods on the NTU RGB+D 60 and 120 datasets. Our model achieved effective results, the accuracy has improved overall and outperformed traditional two-stream networks and temporal-only models. The incorporation of ARN significantly improved the model's ability to differentiate between similar actions.

Table 3 compares the performance of various action recognition models on the NTU-RGB+D 60 and

NTU-RGB+D 120 datasets. The performance of these models is measured by two key accuracy metrics: Cross-Setup and Cross-View. In the context of action recognition, the Cross-Setup metric refers to a training and testing split where actions are captured from different spatial setups (camera locations), while Cross-View focuses on the generalization ability of models across different camera angles.

The ARN-LSTM models evaluated specifically for inward, outward, and inward+outward variants, showcase a substantial leap inaccuracy compared to previous methods. The inwardARN-LSTM variant achieves 97. 1% for Cross-Setup and 97.2% for Cross-View, indicating excellent generalization across setups and views. The outward variant performs slightly lower in Cross-Setup with 94.5% but surpasses the inward variant in Cross-View with an impressive 98.4%. The combined inward+outward ARN-LSTM model achieves the highest accuracy scores of all methods listed, with 99. 1% for Cross-Setup and 99.4% for Cross-View. These results highlight the ARN-LSTM architecture's ability to robustly capture both inward and outward skeletal motion dynamics, leading to near-perfect recognition performance on the NTU-RGB+D 60 dataset.

The ARN-LSTM models also show strong performance on the NTU-RGB+D 120 dataset, with the inward variant scoring 94.9% for Cross-Setup and 91.8% for Cross-View. The outward variant achieves slightly lower accuracies, with 97.7% in Cross-Setup and 94.3% in Cross-View. Finally, the inward+outward variant again outperforms the other methods, achieving a remarkable 99.7% for Cross-Setup and 95.8% for Cross-View. These results indicate that the ARN-LSTM models, especially the combined inward+outward variant, can scale to larger and more diverse datasets without compromising performance.

The ARN-LSTM models, particularly the inward+outward variant, set a new benchmark for performance on both the NTU-RGB+D 60 and NTU-RGB+D 120 datasets, achieving near-perfect accuracy scores. Models such as 4s-shift- GCN, EfficientGCN, and PA-GCN also demonstrate highly competitive performance, though the ARN-LSTM variants slightly outperform them. This progression underscores the importance of architectural innovations, particularly the incorporation of multi-stream, adaptive, and spatio-temporal modeling techniques, in advancing the field of action recognition.

## 5.2 Ablation Study

We conducted an ablation study to better understand each component's contribution by removing different parts of the ARN-LSTM architecture. The results, shown in Table 2, demonstrate that both the multi-stream fusion and ARN components are critical to the model's performance. Removing the ARN block, for example, resulted in a 5% decrease in accuracy, highlighting the importance of attention mechanisms in action recognition.

Table 4 presents our proposed ARN-LSTM approach, a comparative analysis of different fusion feature models for action recognition performance on the NTU-RGB+D 120 dataset using the Cross-View benchmark. The performance metrics are reported for three streams: joint-motion, temporal-motion, and joint-temporal-motion, with accuracy Acc(%), accuracy improvement Acc(↑%), and the most common misclassified actions (Similar Action) for each recognized action. Key observations include the superior performance of the joint-temporal-motion fusion model, which consistently improves recognition accuracy for most actions, such as WalkTowards (91.4%) and ShakeHands (83.9%). Additionally, the Acc(↑%) values highlight substantial improvements in inaccuracy compared to individual streams, particularly for actions like KnockOver (15.3%) and TakePhoto (12.5%). However, certain actions, such as WieldKnife and HitWithObject, remain challenging, with relatively lower accuracies and frequent confusion with similar actions like ShootWithGun and Punch/Slap, respectively. The results underscore the effectiveness of multi-stream fusion in enhancing the accuracy and robustness of action recognition, especially in distinguishing between closely related human activities.

In the temporal-motion stream, WalkTowards achieves the highest accuracy (95.2%), followed closely by Shake-Hands and Hugging (95.2% and 78.8%, respectively), with notable improvements in accuracy for most actions. The joint-temporal-motion fusion consistently provides enhanced recognition performance across most actions. For instance, WalkTowards achieves an accuracy of 98.4%, Hugging improves to 89. 1%, and ShakeHands reaches 95.8%. Several actions, such as Pushing and StepOnFoot, also show substantial improvements in accuracy (up to 9.3% for Pushing). Acc(↑%) values indicate significant improvements in recognition accuracy when utilizing joint-temporal-motion fusion. For example, actions like Pushing, Kicking, and TouchPocket benefit from substantial performance boosts, suggesting that combining joint and temporal information greatly aids in disambiguating these actions. The similar action column highlights the challenges of the task, with certain actions frequently being confused with others. For example, Punch/slap is often misclassified as HitWithObject across all streams, while StepOnFoot is frequently confused with Kicking.

In summary, our proposed ARN-LSTM approach with the joint-temporal-motion fusion consistently outperforms both individual streams (joint-motion and temporal-motion), achieving superior accuracy and minimizing confusion between similar actions. The results indicate that integrating multiple information streams is crucial for improving recognition accuracy, particularly for complex and similar human actions.

Table 4 Comparison of different fusion feature model recognition performance on NTU RGB+D 120 with the Cross-View benchmark.

| Actions | Joint+motion | | Temporal+motion | | | Joint+temporal+motion | | |
|---|---|---|---|---|---|---|---|---|
| | Acc(%) | Similar Action | Acc(%) | Acc↑↓(%) | Similar Action | Acc(%) | Acc↑↓(%) | Similar Action |
| Punch/slap | 49.1 | HitWithObject | 54.1 | ↑5 | HitWithObject | 60.3 | ↑6.2 | HitWithObject |
| Kicking | 70.1 | StepOnFoot | 72.9 | ↑1.8 | StepOnFoot | 74.9 | ↑2.0 | StepOnFoot |
| Pushing | 67.9 | KnockOver | 70.8 | ↑2.9 | KnockOver | 80.3 | ↑9.4 | KnockOver |
| Pat on the back | 58.1 | Touch pocket | 71.2 | ↑13.1 | Touch pocket | 84.1 | ↑12.9 | Touch pocket |
| Point finger | 59.9 | RockPaperScissors | 63.9 | ↑4.0 | RockPaperScissors | 78.1 | ↑14.2 | RockPaperScissors |
| Hugging | 86.9 | High-five | 90.6 | ↑3.7 | High-five | 88.3 | ↓2.3 | High-five |
| Giving object | 51.6 | GrabStuff | 49.8 | ↓1.8 | GrabStuff | 59.7 | ↑9.9 | GrabStuff |
| Touch pocket | 57.3 | Pat on the back | 55.2 | ↓2.1 | Pat on the back | 57.3 | ↑2.1 | Pat on the back |
| Shake hands | 64.2 | Cheers | 58.9 | ↓5.3 | Cheers | 73.4 | ↑14.5 | Cheers |
| Walk towards | 91.1 | Walk apart | 89.9 | ↓1.2 | Walk apart | 91.4 | ↑1.5 | Walk apart |
| Walk apart | 86.8 | Walk towards | 92.4 | ↑5.6 | Walk towards | 90.9 | ↓1.5 | Walk towards |
| HitWithObject | 30.5 | Punch/slap | 45.4 | ↑14.9 | Punch/slap | 45.1 | ↓0.3 | Punch/slap |
| WieldKnife | 43.9 | ShootWithGun | 35.2 | ↓8.7 | ShootWithGun | 43.3 | ↑8.1 | ShootWithGun |
| KnockOver | 65.1 | Pushing | 66.0 | ↑0.9 | Pushing | 81.3 | ↑15.3 | Pushing |
| GrabStuff | 53.8 | CarryObject | 64.0 | ↑10.2 | CarryObject | 72.4 | ↑8.4 | CarryObject |
| ShootWithGun | 43.5 | WieldKnife | 46.9 | ↑3.4 | WieldKnife | 63.9 | ↑17.0 | WieldKnife |
| StepOnFoot | 65.2 | Kicking | 73.9 | ↑8.7 | Kicking | 79.2 | ↑5.3 | Kicking |
| High-five | 71.9 | Hugging | 79.7 | ↑7.8 | Hugging | 90.2 | ↑10.5 | Hugging |
| CheersAndDrink | 63.8 | Shake hands | 64.0 | ↑0.2 | Shake hands | 83.9 | ↑19.9 | Shake hands |
| CarryObject | 84.3 | GrabStuff | 89.6 | ↑5.3 | GrabStuff | 89.6 | -- | GrabStuff |
| TakePhoto | 56.8 | ShootWithGun | 76.0 | ↑19.2 | ShootWithGun | 63.5 | ↓12.5 | ShootWithGun |
| Follow | 55.9 | SupportSomebody | 80.5 | ↑24.6 | SupportSomebody | 81.1 | ↑0.6 | SupportSomebody |
| Whisper | 68.0 | SupportSomebody | 74.1 | ↑6.1 | SupportSomebody | 72.7 | ↓1.4 | SupportSomebody |
| ExchangeThings | 61.4 | Giving object | 76.8 | ↑15.4 | Giving object | 74.0 | ↓2.8 | Giving object |
| SupportSomebody | 66.7 | Whisper | 83.3 | ↑16.6 | Whisper | 83.4 | ↑0.1 | Whisper |
| RockPaperScissors | 64.4 | Point finger | 81.3 | ↑16.9 | Point finger | 75.6 | ↓5.7 | Point finger |

**Acc:** The accuracy of the model in recognizing the action.
**Acc↑:** The improvement (if any) in accuracy when using joint-temporal-motion fusion compared to individual streams.
**Acc↓:** The descend (if any) in accuracy when using joint-temporal-motion fusion compared to individual streams.
**Similar Action:** The most frequent misclassified or confused action.

In Figure 5, the confusion matrices illustrate the performance of the ARN-LSTM method using joint, temporal, and motion features on the NTU RGB+D 120 dataset with the cross-view and cross-subject. The diagonal dominance in both matrices indicates a high accuracy across most of the 26 action classes, such as "Kicking," "Pushing," and "Support somebody," where the model consistently makes correct predictions. However, there are notable mis-classifications, particularly between actions with similar motion dynamics, such as "Punch/slap" and "Hit with object," which are frequently confused in both settings. These results suggest that while the system effectively distinguishes many actions, it faces challenges in separating actions with overlapping visual or motion characteristics, especially when there are variations in viewpoint or subject. Despite these challenges, the overall performance remains strong, as evidenced by the large number of correct classifications in both cross-view and cross-subject experiments.

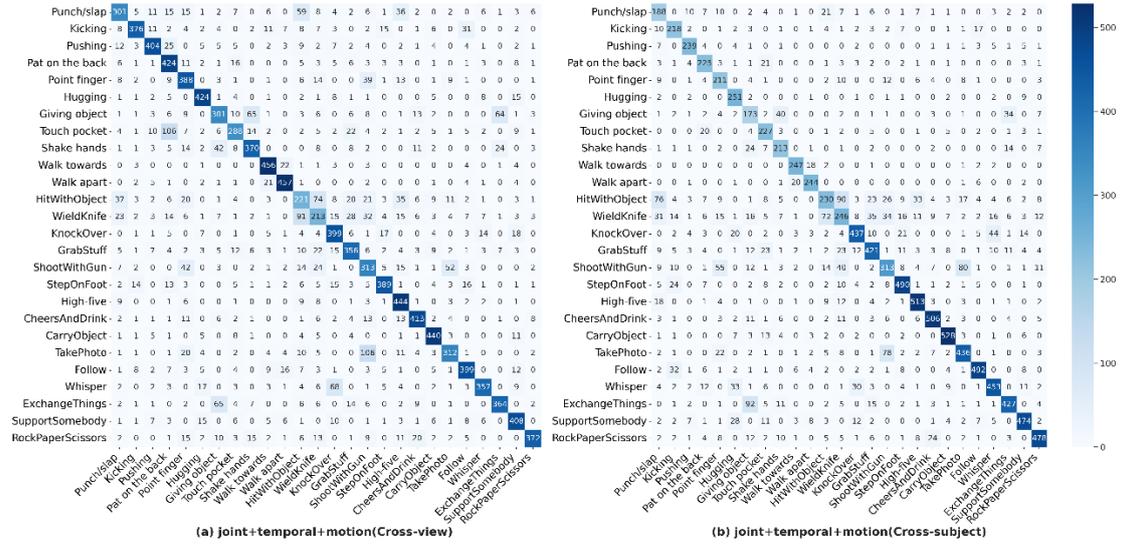

Figure 5 Confusion matrix of ARN-LSTM with joint, temporal and motion features on the NTU RGB+D 120 dataset.

These confusions experiments presented above are logical and reliable, as they rely solely on the coordination of body parts for information, and these interactions are expected to mimic human movements (e.g. by reaching out to one another, primarily identifiable by the item (or its nonexistence) each person possesses. Our view is that this robustly supports the use of RGB visual data for enhanced interaction recognition. A more detailed examination of the performance of ARN−LSTM is shown in Figure 6, which contains the accuracy per interaction class for both cross-subject and cross-view benchmarks. Some interactions are significantly more challenging than others, with a recognition rate much inferior than the average.

6 Conclusion

In this paper, we introduced ARN-LSTM, a novel multi-stream model for action recognition that effectively combines joint, motion, and temporal features. The model can capture complex spatial-temporal dependencies by integrating Attention Relation Networks and a TD-LSTM. The proposed ARN-LSTM approach is a sophisticated neural network model adept at processing diverse forms of object data, including federated streams, joint streams, temporal streams, and general relational data. It leverages relational modeling and attention mechanisms to enhance model performance. The flexible parameter configuration facilitates the implementation of various network architectures. On the two large-scale datasets, NTU RGB+D 60 and NTU RGB+D 120 dataset, the proposed ARN-LSTM leads to effective performance on group activity recognition tasks, with the average rate of accuracy reaching 96%. However, we believe there is still potential for improvement. Future work will focus on extending the model to handle larger datasets and exploring its application to real-time action recognition systems. Inspired by[10][66][67], we will also attempt to use higher-level information to improve the results, such as the Grap Convolutional Neural Network method.

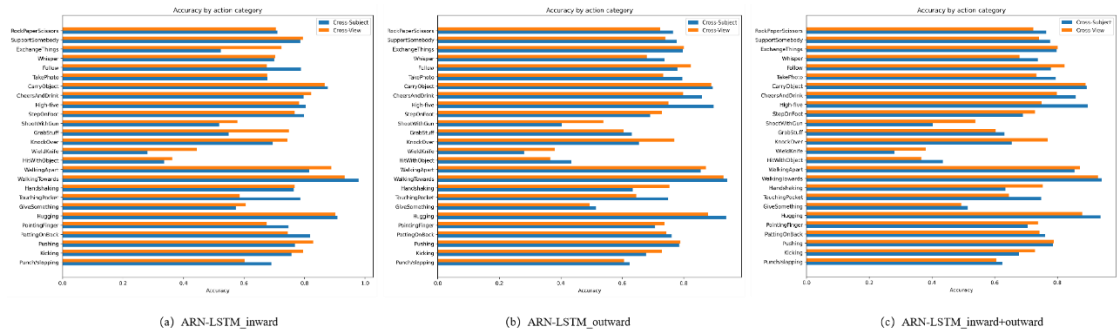

Figure 6 Performance per interaction class of proposed ARN-LSTM method on the NTU GRB+D 120 dataset(total 26 mutual classes), with the Crosssubject(the blue bar) and CrossView(the orange bar) benchmarks. (a)presents the ARN-LSTM_inward approach's accuracy in each class.(b) presents the ARN-LSTM_outward approach's accuracy in each class. (c) presents the AR N-LSTM_inward+ outward approach's accuracy in each class.

**Acknowledgments**
This work was partly supported by the University Characteristic Innovation Projects of Guangdong Province of China (Grant: 2023KTSCX196). The authors would like to thank the anonymous referees for their insightful comments, which greatly improved the quality of this paper.

**Author contributions**

Chuanchuan Wang wrote the main manuscript text and constructed the main idea of the proposed framework. Ahmad Sufril Azlan Mohamed reviewed the manuscript and gave suggestions for writing and publishing. Xiang Li prepared the first results of the framework and helped develop the proposed framework. Xiao Yang helped to preprocess the results of the experiment. All of the authors edited the submitted version of the manuscript.

**Competing interests**
The authors declare that they have no known conflict of competing interests, this work is only for study.

**Data Availability**
The datasets used for the study can be available at https://doi.org/10.6084/m9.figshare.27427188.v1. The raw dataset from the dataset home page https://rose1.ntu.edu.sg/dataset/actionRecognition/.